# Incept-N: A Convolutional Neural Network based Classification Approach for Predicting Nationality from Facial Features

Masum Shah Junayed, Afsana Ahsan Jeny, Nafis Neehal*
Department of Computer Science & Engineering
Daffodil International University
Dhaka, Bangladesh
E-mail: {junayed15-5008, ahsan15-5278, nafis.cse}@diu.edu.bd

*Abstract*— Nationality of a human being is a well-known identifying characteristic used for every major authentication purpose in every country. Albeit advances in application of Artificial Intelligence and Computer Vision in different aspects, its' contribution to this specific security procedure is yet to be cultivated. With a goal to successfully applying computer vision techniques to predict a human's nationality based on his facial features, we have proposed this novel method and have achieved an average of 93.6% accuracy with very low misclassification rate.

## I. INTRODUCTION

Facial recognition is a complicated process that involves using facts and experience to set an average face to measure other faces too. The capability to identify faces is very significant to many aspects of life. It not only helps us to identify those close to us but also approves us to recognize individuals we do not know so that we can be more conscious of probable dangers.

The human face is an extremely rich inspiration that gives amazing information for adaptive social interaction with humans. Over the past few decades, a lot of attempts has been devoted to the biological, psychological, and cognitive sciences areas, to detect how the human brain perceives, describes, and remembers faces [11].

The human face is a complex visual pattern that along with general categorical information as well as eccentric, identify specific, primary information. By this categorical information, we mean that some aspects of a face are not appointed to that individual face but are shared by subsets of faces. These aspects can be used to impose both familiar and unfamiliar faces to general semantic groups such as district or nationality.

With the development of computer technology and digital image processing technology, people began to explore the method of automatic nationality identification by a computer, this method is mainly based on the distances, angles, and areas and other features of the people to calculate the similarity between the human images and then determine the people [5]. The most ancient method of nationality identification is to notice the living habits, morphological structure and other features of persons. This classification method is fully artificial, the workload is massive, and need the professional staffs who have a wealth of professional knowledge and experience to guide.

In this paper, we use the transfer learning technique to retrain the Inception-v3 [8] model of Tensor Flow [1] on the dataset of 5 countries (China, Germany, India, Jamaica, and Zimbabwe). We fulfilled an efficient national identity model using a short training time and obtain a higher accuracy. The remaining paper is arranged in the following manner: Details of Convolutional Neural Network (CNN) and Inception-v3 [8] model are discussed in Section II. The comparison with other papers is discussed in Section III. Data collection and training are discussed in Section IV. Performance analysis is done in Section V. Finally conclusion with some future work scopes is described in Section VI and Section VII.

## II. BACKGROUND STUDY

This experiment is based on the Inception-v3 [8] model of TensorFlow [1] platform and also used CNN [2].

TensorFlow [1] as the second generation of Google artificial intelligence learning system has got much interesting and representation in the field of machine learning in all over the world. TensorFlow [1] has ranked first in all deep learning and machine learning programs so far. TensorFlow [1] has the benefits of high suitability and high facility, and with the help of TensorFlow researchers, the ability of TensorFlow is developed. Today, Google has opened a number of trained models on the TensorFlow's official website, to simplify the use of researchers in different sectors.

Inception-v3 [8] is one of the trained models on the TensorFlow [1]. It is a rethinking for the initial structure of computer vision after Inception-v1 [9], Inception-v2 [9] in 2015. The Inception-v3 [8] model is trained on the ImageNet datasets, containing the information that can identify 1000 classes in ImageNet. Inception-v3 [8] consists of two parts: Feature extraction part with a convolutional neural network (CNN) and Classification part with fully-connected and softmax layers [10].

Convolutional Neural Networks (ConvNets or CNN) are a class of Neural Networks that have vindicated very effectively in areas such as image classification and recognition. ConvNets have been effective in identifying objects, faces, and traffic. Typically three main types of layers are used to build ConvNet architectures:

- Convolutional Layer,
- Pooling Layer and
- Fully-Connected Layer.

The first layers of a CNN [2] strain (large) features that can be acknowledged and illustrated relatively easy. As a result of convolution in neuronal networks, the image is split into perceptrons, creating local receptive fields and finally compressing the perceptrons in feature maps of size $m_2 \times m_3$.

Thus, this map stores the information where the feature occurs in the image and how well it corresponds to the filter. Hence, each filter is trained spatial in regard to the position in the volume it is applied to [2]. In each layer, there is a bank of m1 filters. The number of how many filters are applied in one stage is equivalent to the depth of the volume of output feature maps. Each filter detects a particular feature at every location on the input. The output $Y_i^{(l)}$ of layer l consists of $m_1^{(l)}$ feature maps of size $m_2^{(l)} \times m_3^{(l)}$. The $i^{th}$ feature map, denoted $Y_i^{(l)}$, is computed as

$$Y_i^{(l)} = B_i^{(l)} + \sum_{j=1}^{m_1^{(l-1)}} K_{i,j}^{(l)} * Y_j^{(l-1)}$$

where $B_i^{(l)}$ is a bias matrix and $K_i^{(l)}$,j is the filter of size $2h_1^{(l)}+1 \times 2h_2^{(l)}+1$ connecting the $j^{th}$ feature map in layer (l−1) with $i^{th}$ feature map in layer [2]. Later the layers detect increasingly (smaller) features that are more abstract (and are usually present in many of the larger features detected by earlier layers). The pooling layer l has two hyper parameters, the spatial extent of the filter $F^{(l)}$ and the stride $S^{(l)}$. It takes an input volume of size $m_1^{(l-1)} \times m_2^{(l-1)} \times m_3^{(l-1)} \times$ and provides an output volume of size $m_1^{(l)} \times m_2^{(l)} \times m_3^{(l)}$ where;

$$m_1^{(l)} = m_1^{(l-1)}$$
$$m_2^{(l)} = (m_2^{(l-1)} - F^{(l)})/ S^{(l)}$$
$$m_3^{(l)} = (m_3^{(l-1)} - F^{(l)})/ S^{(l)} + 1$$

The last layer of the CNN [2] is able to make an ultra-specific classification by combining all the specific features detected by the previous layers in the input data. It also has a certain degree of translation, rotation and distortion invariance of the image. It has made great progress in the field of image classification [2].

If l−1 is a fully connected layer;

$$y_i^{(l)} = f(z_i^{(l)}) \text{ with } z_i^{(l)} = \sum_{j=1}^{m_1^{(l-1)}} w_{i,j}^{(l)} y_i^{(l-1)}$$

Otherwise;

$$y_i^{(l)} = f(z_i^{(l)}) \text{ with } z_i^{(l)}$$

$$= \sum_{j=1}^{m_1^{(l-1)}} \sum_{r=1}^{m_2^{(l-1)}} \sum_{s=1}^{m_3^{(l-1)}} W_{i,j,r,s}^{(l)} (Y_i^{(l-1)}) \, r_{l^s}$$

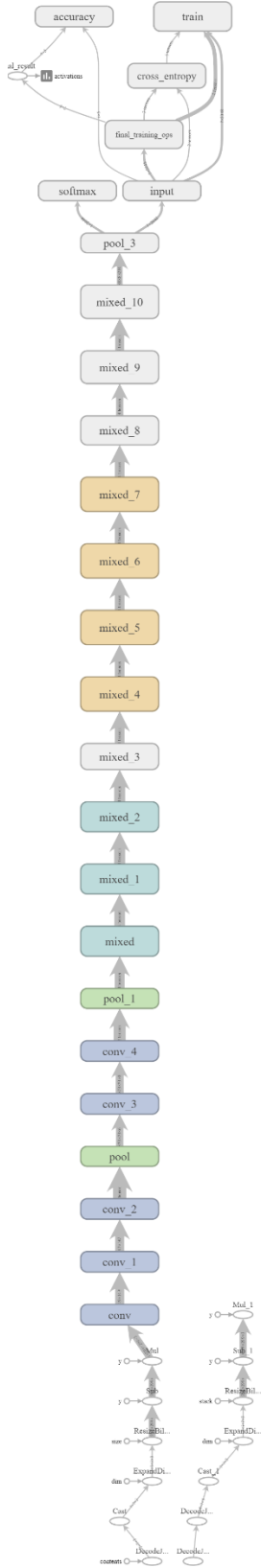

Figure 1. Main graph of Inception v3 model.

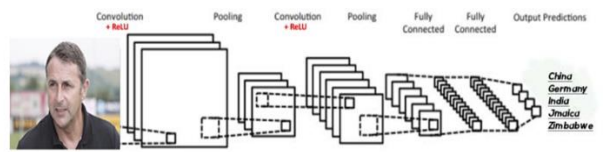

Figure 2. Structure of Convolutional Neural Network (CNN).

*Declaration – This paper is currently under review in "3rd IEEE International Conference on Image, Vision and Computing (ICIVC) 2018, China".

TensorFlow [1] makes available in detailed tutorials for us to retrain Inception's final Layer for new categories using transfer learning. Transfer learning is a new machine learning way which can use the existing knowledge learned from one environment and find an answer to the other new problem which is different but has some relation to the old problem. Measured with the traditional neural network, it only needs to use a small amount of data to train the model, and achieve high accuracy with a short training time [5] [7].

### III. LITERATURE REVIEW

Inception v3 [8] model used in many experiments. Among them:

In 2017, Xiaoling Xia and Cui Xu from College of Computer Science, Donghua University used the transfer learning technique to retrain the Inception-v3 [8] model of TensorFlow [1] on the flower category datasets [11] [13] of Oxford-I7 and Oxford-102 for Flower Classification. The classification accuracy of the model was 95% on Oxford-I7 flower dataset and 94% on Oxford-102 flower dataset [5].

In 2017, Alwyn Mathew*a, Jimson Mathewa, Mahesh Govindb, Asif Mooppanb from bVuelogix Technologies Pvt Ltd used Google's TensorFlow[1] deep learning a framework to train, validate and test the network for Intrusion Detection and the accuracy was 95.3%. But the proposed network is found to be harder to train due to vanishing gradient [3] and degradation problems [3].

In 2017, Brady Kieffer1, Morteza Babaie2 Shivam Kalra1, and H.R.Tizhoosh1 used CNN and Inception v3 [8] model for Histopathology Image Classification [6]. All experiments are done on Kimia Path24 dataset and the accuracy was 56.98% [6]. In 2017, Xiao-Ling Xia 1, Cui Xu*2, Bing Nan3 worked for Facial Expression Recognition based on the Inception-v3 [8] model of TensorFlow [1] platform. They used CK+ dataset [15] and selected 1004 images of facial expression. Their accuracy was 97% but it wasn't based on dynamic sequences [7].

In 2016, Bat-Erdene.B and Ganbat.Ts worked on Effective Computer Model for Recognizing Nationality from Frontal Image [4]. They used SVM, AAM, ASM and the accuracy was 86.4%. Their experiment was worked manually and images must be the frontal face image that has smooth lighting and does not have any rotation angle.

Our experiment is based on the Inception-v3 [8] model of TensorFlow [1] platform for Nationality Recognition based on facial features with Deep Learning. Nobody did it before. This is the first approach from us. It is worked automatically and images have rotation angle and translation.

### IV. METHODOLOGY

In this section, the following part is as follows: first we make a flowchart [12] of our experiment; second, we provide a simple introduction on the dataset; third, we give about the data preprocessing; then, we discuss the model installation; finally, we introduce about the train model.

A flowchart [12] is a type of diagram that represents a workflow or process.

The flowchart [12] shows the steps of boxes, and their order by connecting the boxes with arrows. Flowcharts [12] are used in analyzing, designing or managing a process. The following diagrammatic representation illustrates a solution model to our system.

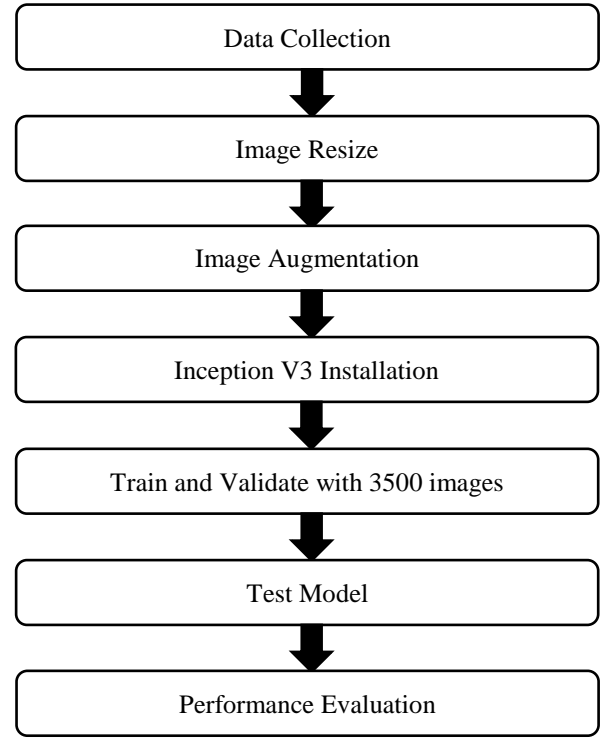

Figure 3. Flowchart of the system model.

#### I. Dataset

There are many countries in this world and also many people. There is a similarity in the appearance of human faces. For recognizing nationality, we have collected 600 images of five countries for our experiment. They are China, Germany, India, Jamaica, and Zimbabwe.

#### II. Data preprocessing

In order to promote the effect of image classification, image preprocessing is a very significant stage. The learning method of convolution neural network belongs to observe and direct the execution of our activity in machine learning, so in the image preprocessing step we need to label the data. Then we have resized the data and also augmented (Rotate +30, Rotate -30, Translation, Lighting and Flip). Finally, we have found 3600 images for training.

#### III. Model installation

This experiment is based on the Inception-v3 [3] model of TensorFlow [1] platform. The processor is 2GHz Intel i3, memory 4GB 1600MHz DDR3, System type: 64-bit Operating System, x-64 based processor.

First of all, we have downloaded TensorFlow [1]. Then we have downloaded Inception v3 [8] model. We have also used the transfer learning method which keeps the parameters of the prior layer and have removed the final layer of the Inception-v3 [8] model, then retrain a final layer.

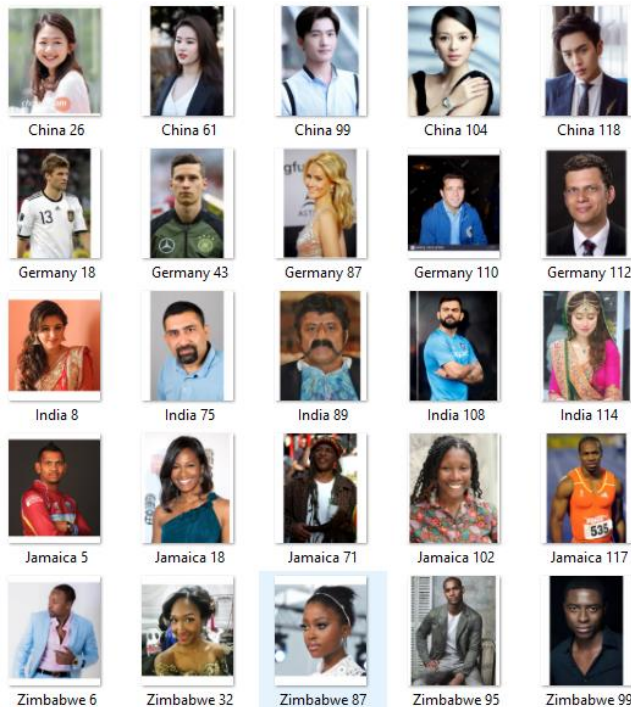

Figure 4. The example of our dataset.

### IV. Train model

In this step, we should keep the parameters of the previous layer, then remove the final layer and input our dataset to retrain the new last layer.

The last layer of the model is trained by back propagation algorithm, and the cross-entropy cost function is used to synthesize the weight parameter by calculating the error between the output of the softmax layer and the label vector of the given test category [5] [7].

We have also created Confusion Matrix for final accuracy. From Confusion Matrix, we have calculated Precision, Recall, Accuracy, and F1-Score. And finally, we have calculated Macro Average Accuracy of our experiment.

Here is the Confusion Matrix of our model. From the following Confusion matrix of Table I, we can tell that our model has given a very high number of True Positive values.

TABLE I. CONFUSION MATRIX

|  | Chia | Germany | India | Jamaica | Zimbabwe |
|---|---|---|---|---|---|
| China | 18 | 1 | 0 | 1 | 0 |
| Germany | 0 | 19 | 1 | 0 | 0 |
| India | 1 | 0 | 16 | 1 | 2 |
| Jamaica | 0 | 1 | 0 | 16 | 3 |
| Zimbabwe | 1 | 1 | 0 | 3 | 15 |

### V. RESULT ANALYSIS

Figure 5 and figure 6 show the variation in accuracy and cross-entropy based on our training dataset. The orange line represents the training set, and the blue line represents the validation set.

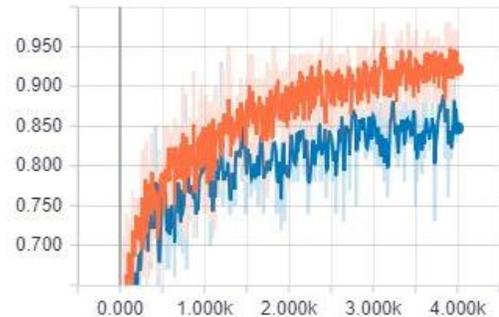

Figure 5. The variation of accuracy on the training dataset.

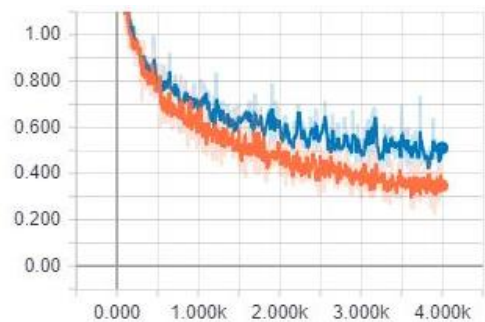

Figure 6. The variation of cross entropy on the training dataset.

TABLE II. DESCRIPTION OF THE TWO FIGURES

| Dataset | Index | Performance |
|---|---|---|
| Dataset | the accuracy of the training set | 95% |
|  | the accuracy of the validation set | 89%-90% |
|  | the cross-entropy of the training set | 0.24 |
|  | the cross-entropy of the validation set | 0.41 |

Table II shows the description of the two figures. For our dataset, the training accuracy can reach to 95%, and the validation accuracy can be maintained at 89% -90%.

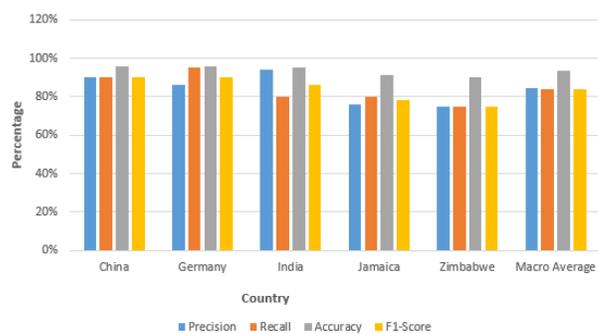

Figure 7: Precision, Recall, Accuracy and F1-Score graph.

Figure 7 shows the precision, recall, accuracy and F1-Score graph of China, Germany, India, Jamaica, and Zimbabwe and also show the precision, recall, accuracy and F1-Score of Macro-Average [14] [15].

TABLE III. THE ACCURACY OF FIVE COUNTRIES AND FINAL ACCURACY

| Country | Accuracy |
|---|---|
| China | 96% |
| Germany | 96% |
| India | 95% |
| Jamaica | 91% |
| Zimbabwe | 90% |
| **Macro Average** | **93.6%** |

Table III shows the accuracy of five countries from the graph. For our dataset, the accuracy of China is 96%, Germany is 96%, India is 95%, Jamaica is 91%, Zimbabwe is 90% and the final accuracy is 93.6%.

## VI. FUTURE WORK

Since the Inception-v3 [3] model of TensorFlow[1] platform is generated by Google [10] and we have used it. So our future work is to study and develop a more effective model so that we can use that model and can increase our accuracy.

## VII. CONCLUSION

In this paper, based on the Inception-v3 model of TensorFlow[1] platform, we use the transfer learning technology to identify the nationality of five countries based on our dataset. And we get the accuracy of the model is 93.6%. Hopefully, in near future, we can improve this method and achieve better accuracy.